\documentclass{article}

\usepackage[english]{babel}

\usepackage[letterpaper,top=2cm,bottom=2cm,left=3cm,right=3cm,marginparwidth=1.75cm]{geometry}

\usepackage{amsmath,amsfonts,amssymb}
\usepackage{bm}
\usepackage{graphicx}
\usepackage{float}
\usepackage[colorlinks=true, allcolors=blue]{hyperref}
\usepackage{authblk}
\usepackage{makecell}

\title{Silver medal Solution for Image Matching Challenge 2024}
\author{Yian Wang}
\affil{North Broward Preparatory School, Building 800,7600 Lyons Road, Coconut Creek, 33073, FL, USA}
\date{}

\begin{document}
\maketitle

\begin{abstract}
Image Matching Challenge 2024 is a competition that aims to build 3D maps from sets of images taken in different scenes and environments. The competition requires participants to solve the fundamental computer vision problem of image matching, including dealing with complex factors such as different viewing angles, lighting and seasonal changes. The diversity of image shooting angles and environments places higher demands on the robustness and diversity of the participating algorithms. There are many challenges in image matching, such as differences in surface texture and surroundings that may lead to significant degradation in algorithm performance. Especially when matching images taken from different viewpoints, traditional methods are difficult to cope with these variations, so new methods need to be developed to improve the accuracy and stability of matching. This project develops a Pipeline method, which consists of the following steps: firstly, image data features are extracted using the pre-trained EfficientNet-B7 model, and similar image pairs in the image set are filtered and sorted according to the cosine distance. Then, two key point feature extraction methods, KeyNetAffNetHardNet and SuperPoint, are used to obtain feature point locations. Next, AdaLAM and SuperGlue are used for keypoint matching. Finally, 3D spatial location relations are performed by Pycolmap to determine the final matching results of image pairs. The methodology of this project has achieved an excellent score on the leaderboard of Image Matching Challenge 2024. The project scored 0.167 on the private leaderboard and performed well on the public test set. The experimental results show that the combination of KeyNetAffNetHardNet and SuperPoint has a significant advantage in keypoint detection and matching. The project conducted several experiments to verify the performance of different feature extraction and matching methods and found that the combination of KeyNetAffNetHardNet and SuperPoint worked best.
\end{abstract}



\section{Introduction}
\label{sec:sample1}

Image Matching Challenge 2024 [1] The aim of the competition is to construct 3D maps using sets of images from different scenarios, environments, and domains, such as drone shots, deep in dense forests, and nighttime shots.

Compared to last year's Image Matching Challenge 2023 [2], the requirements are very high, such as model diversity, variability, and robustness. Because each photo is taken from a slightly different angle, the shadows may also vary depending on the time of day and season in which the photo was taken. One photo may have been taken from the ground, another from a step, and yet another from a drone. Matching images from different viewpoints is a fundamental computer vision problem that has not yet been fully solved. Factors such as surface texture or surroundings can cause performance degradation in an otherwise well-performing algorithm. The project developed a pipeline approach , the specific process is that the image set of each scene The image data features are first extracted using ImageNet weights from the pre-training model efficientnet-b7 [3], filtered based on the cosine distance, and the first n image pairs of the image set are sorted according to their similarity. Then the retrieved image pairs one by one use two keypoint feature detectors to extract the relevant feature point locations, use two keypoint matching algorithms to match all the matched point locations for the matching calculation, and save the matched pairs (match pairs) successfully. Finally, the image pairs of matched successful points (match pairs) are integrated to exclude the same and then superimposed into pycolmap to calculate the final 3d spatial positional relationship (position and pose estimation). The score of this pipeline is 0.168089 in leaderboard's private list test and public test.

previous study
In order to complete the Image Matching Challenge 2024, this project refers to the work of Image Matching Challenge 2023 or 2022, many researchers have disclosed their own use of the method, most of them are divided into the following four parts of image retrieval, feature extraction, matching, 3D reconstruction are implemented separately, this This project introduces the general method of these four parts

\subsection{Image Retrieval}
Image retrieval is a key task in computer vision aimed at identifying highly similar image pairs from large-scale image sets. In recent years, with the advancement of deep learning, most image retrieval methods now rely on deep learning techniques to extract global features of an image. These methods mainly use convolutional neural networks (CNNs) or transformer networks.

\textbf{Convolutional Neural Networks (CNNs)} are widely used for image retrieval due to their ability to efficiently capture spatial hierarchies in images.NetVLAD [4] is a notable CNN-based approach that combines the benefits of the VLAD ( Vector Local Aggregation Descriptor ) aggregation layer with deep learning features to significantly improve retrieval performance on a variety of datasets. EfficientNet is another noteworthy CNN architecture that balances the depth, width, and resolution of the network to achieve state-of-the-art results with fewer parameters and computational resources.ConvNeXt [5] is a modernised CNN architecture that combines the design elements of transformer [6,7] networks to further enhance its retrieval capabilities.

\textbf{Transformer networks} have received increasing attention for their effectiveness in capturing long-range dependencies and contextual information in images.DINOv2 [8] is a self-supervised transformer-based model that learns robust visual representations without the need for labelled data and performs well in image retrieval tasks.CLIP [9] and its variants EVA-CLIP [10] exploits large-scale image-text pair pre-training to achieve zero-sample migration learning, and achieves significant results in image retrieval benchmarks.ViT [11] introduces a visual transformer architecture and applies the transformer model directly to an image block, setting new records in a variety of visual tasks, including image retrieval.

\subsection{Feature extraction}
With the development of SIFT [12], local features have become an important part of computer vision.The classical SIFT-based approach involves three steps: keypoint detection, orientation estimation, and descriptor extraction. Other method AKAZE [13], which is implemented using a combination of fast significant diffusion and nonlinear scale space.

Recently descriptor extraction usually trains deep networks on patches, usually from SIFT keypoints. They include L2-Net [14], HardNet [15], TFeat [16]. There have also been several attempts by researchers to learn keypoint detectors at descriptors alone, including TCDet [17], Keynet [18]. There is also another end-to-end training method that contains SuperPoint [19],D2-Net [20],R2D2 [21],ALIKED [22].

\subsection{Match}
Local feature matching is usually done by examining keypoints, computing visual descriptors, and using Nearest Neighbor (NN) [23] to search for matches to these features, filtering out incorrect matches. Others are finding matches using robust solvers such as RANSAC [24].
Recent approaches based on deep learning matching, such as SuperGlue [25], LightGlue [26], the former based on a flexible context aggregation mechanism and graph structure based on attention, and the latter based on transformer as well as self-attention, cross-attention.
\subsection{Rebuild}
The most popular frameworks in SFM are VisualSFM [27] and COLMAP [28], the latter being commonly used for projects that require it as it produces ground truth and also serves as a backbone for multi-view tasks.

\section{Methodologies}
In this section, the overall architecture of the pipeline of this project is described in detail, as shown in Figure xx, the pipeline contains five parts: image retrieval, keypoint feature extraction, keypoint feature matching, integration, and spatial location estimation, which are described in detail in the following sections

\subsection{Image Retrieval}
The project filters similar image pairs in the dataset for each scene. It uses an efficient convolutional neural network model ( EfficientNet-B7 ) to extract the global descriptors of the images and calculates the similarity between the images based on these descriptors. The project has set the N value to 45, how the total number of images in the dataset of the scenario is smaller than this, the project will use an exhaustive search method to obtain all possible pairs of images. Instead the Euclidean distance between global descriptors is obtained using the EfficientNet-B7 model and then similar image pairs are filtered based on the distance matrix and similarity threshold.

\subsection{Key Point Feature Extraction}
Key point feature advancement is in a crucial position in the whole pipeline and the project has tried many key feature extraction methods such as:

ALIKED ,DISK[29],SIFT, SuperPoint ,Dog-hardnet but the effect is not very ideal, after the project many times to try to change the keypoint detection threshold, the number of detected features, the size of the image shape, the integration of multiple models, etc., found that these parameters affect the effect of this model, and ultimately found that the number of detected features is 8081, the keypoint detection threshold is 0.001023349, the image shape size is 1024 when the best results. threshold is 0.001023349 and image shape size is 1024 is the best. The project method uses two key point feature extraction methods, respectively KeyNetAffNetHardNet ,SuperPoint, the next project will introduce these two feature extraction methods

\subsubsection{Key-Aff-HardNet}
Key-Aff-HardNet consists of three modules: direction estimation module, feature point detector module, and descriptor module, where the direction estimation module OriNet [30] is used for angle estimation, the feature key point detector uses KeyNet detector as well as AffNet [31] shape estimation, and the descriptor uses HardNet network.

\textbf{HardNet} is a compact descriptor for local descriptor learning, at the heart of which is a novel loss function designed to maximise the distance between the nearest positive and nearest negative samples in a batch. The loss function is inspired by the SIFT matching criterion and improves the discriminative power of the descriptors by minimising the distance between the matching descriptors and the nearest mismatched descriptors. Specifically, the method first generates a batch containing matched local patches, then computes the descriptors of these patches over the network and constructs a distance matrix. Next, the closest mismatched descriptors are selected from each pair of matched samples to form a triad. Finally, the model is optimised by calculating the triad loss.

\begin{figure}[!h]
    \centering
    \includegraphics[width=\textwidth]{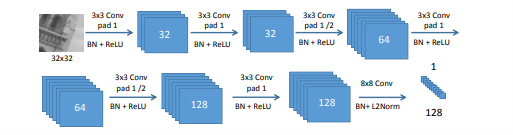}
    \caption{HardNet's network architecture}
    \label{fig2}
\end{figure}

The HardNet architecture employs the L2Net CNN architecture using a series of convolutional and batch normalisation layers to output a 128-dimensional descriptor, and the network structure is illustrated in Fig.\ref{fig2} This study demonstrates that HardNet significantly outperforms hand-crafted and other learnt descriptors in several real-world tasks, including image retrieval and wide-baseline stereo vision matching. The performance improvement is attributed to its simple but effective learning objectives and the full utilisation of existing datasets.

One of the reasons why the HardNet network works better than other models is the use of triplet loss, as the discriminative power of the model is improved by maximising the distance between matched sample pairs and the nearest negative sample. The principle is to generate a set of matched local patch pairs from the training data, compute the Euclidean between all sample pairs to form a distance matrix, and from the distance matrix, select the non-matching samples that are closest to each anchor and positive sample, i.e., the hardest negative sample. For each sample pair, the loss value is calculated based on the distance of the matched sample and the distance of the nearest negative sample, and the loss values of all sample pairs are averaged to obtain the final loss. Equation (1) is given below:

\begin{equation}
L=\frac{1}{n} \sum_{i=1}^n \max \left(0,1+d\left(a_i, p_i\right)-\min \left(d\left(a_i, p_{j_{\min }}\right), d\left(a_{k_{\min }}, p_i\right)\right)\right)
\end{equation}

\noindent Among them:

$n$ is the number of sample pairs in the batch.

$a_i$ and $p_i$ are the anchor and positive samples of the positive sample pair, respectively.

$p_{j_{\min }}$ is the nearest negative sample to the anchor point $a_i$.

$a_{k_{\min }}$ is the nearest negative sample to the positive sample $p_i$.

$d(x,y)$ is the distance between sample $x$ and sample $y$.

$d(a_i, p_i)$ penalises dissimilarity between correct matches, - $\min (d(a_i, p_{j_{\min }} )$, $d(a_{k_{\min }}, p_i))$ penalises similarity between incorrect matches

\textbf{AffNet} is a convolutional neural network (CNN) for estimating local affine shapes, aiming to improve the reliability and accuracy of image feature matching. Traditional feature detectors such as Hessian-Affine do not perform well under illumination and viewpoint changes, and cannot guarantee reliable feature matching.AffNet learns the affine shape and orientation, and proposes a new Hard Negative-Constant Loss function, which combines the advantages of the ternary loss and the contrast loss, to optimise the affine region estimation. The specific formula (2) is as follows:


\begin{equation}
L_{\text{HardNeg}}=\sum_{i=1}^n \max \left(0,1+d\left(s_i, \hat{s}_i\right)-d\left(s_i, N_i\right)\right)
\end{equation}

Equation (2) $d\left(s_i, \dot{s}_i\right)$ indicates the distance between the first $i$ sample and its positive sample. The smaller this distance is, the better it is, as it indicates a higher similarity between the positive samples.

$d\left(s_i, N\right)$ denotes the distance between the $i-t h$ sample and its negative sample. The larger this distance is, the better it is, as it indicates a lower similarity between the negative samples. At the heart of this loss is the comparison of the distance between the positive samples $d\left(s_i, \dot{s}_i\right)$ and the distance between the negative samples $d\left(s_i, N\right)$. For each sample $s_{i}$ the value of
$1+d\left(s_i, \dot{s}_i\right)-d\left(s_i, N\right)$ greater than 0, it is included in the loss, otherwise it is 0.

\textbf{KeyNet} combines hand-crafted and learned Convolutional Neural Network (CNN) filters for efficient and stable keypoint detection in shallow multi-scale architectures. The handmade filters provide the anchor structure, the learned filters are responsible for locating, scoring and ordering the keypoints, and the network extracts the keypoints at different levels through a scale space representation.

\subsubsection{SuperPoint}

SuperPoint is a self-supervised learning framework for interest point detection and description, see Fig. xx. Its approach is based on a fully convolutional neural network architecture and contains a shared encoder and two decoder heads for interest point detection and descriptor generation, respectively. The shared encoder is in VGG style, which reduces the spatial dimension of the input image through convolutional layers, pooling layers and activation functions. The input image is changed from $I\in \mathbb{R}^{H\times W}$ to $\mathbb{R}^{H_{c}\times W_{c}}$ where $H_{c}=H/8$ and $W_c = W/8$, the feature map is reduced to 8 times of the original one. the feature map is mapped to $X \in \mathbb{R}^{H_c \times W_c \times 65}$ in the interest point detection decoder, where 65 channels are for an 8*8 grid region plus an interest point score. The output vector is turned into
 $\mathbb{R}^{H\times W}$ by softmax.

The descriptor decoder maps the feature map to $D \in \mathbb{R}^{H_c \times W_c \times D}$ original image resolution using bicubic linear interpolation, and then compresses the descriptors to unit values via L2 normalisation.SuperPoint used a pre-trained model for inference in this project.

\begin{figure}[!h]
    \centering
    \includegraphics[width=0.8\textwidth]{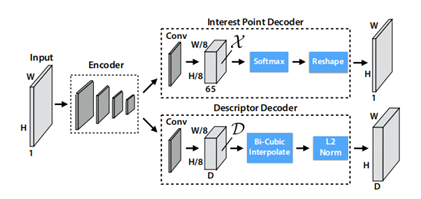}
    \caption{SuperPoint Network Architecture}
    \label{fig3}
\end{figure}

\subsection{Keypoint Feature Matching}
Keypoint Feature Matching The project tried three methods such as LightGlue , SuperGlue, AdaLAM [32].

\subsubsection{AdaLAM}
AdaLAM is an efficient outlier detection method for image matching based on local affine motion verification with sample adaptive thresholding. The method incorporates best practices developed over the years in the field of computer vision, proposes a hierarchical pipeline for efficient outlier filtering, and integrates the utilisation of modern parallel hardware to enable it to process a large number of image keypoint matches in a short period of time. Specifically, AdaLAM filters out false matches by selecting high-confidence and well-distributed seed points and verifying the local affine consistency in their neighbourhood. Its adaptive thresholding mechanism is adjusted by statistical significance test, which improves the algorithm's generalisation ability and robustness in different scenarios. Experimental results show that AdaLAM performs well in several indoor and outdoor scenes and significantly outperforms existing state-of-the-art methods.

\subsubsection{SuperGlue}
SuperGlue is a neural network for image feature matching that solves the feature matching problem between images by combining graph neural networks and attention mechanisms. It focuses on estimating the matching relationship by solving a microscopic optimal transport problem and predicting the matching cost using a graph neural network.SuperGlue introduces self-attention and cross-attention mechanisms that allow it to aggregate contextual information within and between images, and to reason about 3D scenes and feature assignments simultaneously. The method consists of two main modules: an attention graph neural network and an optimal matching layer. The former aggregates image features through self-attention and cross-attention mechanisms, while the latter generates a partial allocation matrix through the optimal transport problem to efficiently deal with occlusions and non-repeatable keypoints.SuperGlue is capable of running in real-time and can be seamlessly integrated into modern Structured Light (SfM) or Just-in-Time Localisation and Map Building (SLAM) systems, providing them with high-quality feature matching and attitude estimation.

\subsection{Integration}
Integration learning is a technique that improves overall predictive performance and robustness by combining multiple weaker base models. Integration methods are particularly effective in the fields of image matching and feature extraction . In this project, the project used two different feature extraction algorithms, SuperPoint and KeyNetAffNetHardNet, both of which can identify key points in an image. And for keypoint matching, the project used two matching techniques, AdaLAM and SuperGlue. In this way, each algorithm can play a role in its area of expertise, while enhancing the performance and adaptability of the whole system by complementing each other. This integration strategy not only optimises the detection and matching process at critical points, but also improves the accuracy and reliability of the model in dealing with complex or changing environments.

\subsection{Estimation of spatial position}
For the spatial position estimation module the project uses the pycolmap library, which is an interface library for Python users that allows developers to use the functionality of COLMAP directly in the Python environment. COLMAP is a state-of-the-art computer vision software that focuses on 3D reconstruction of unordered image collections. Using pycolmap, users can easily implement spatial position estimation of images.

\section{Experiments and the Results}
\subsection{Leaderboard Submission}
The project found this idea in last year's image matching adjustment in 2023, and the project reproduced it and applied it to the image matching challenge in 2024. The project also tried many experiments and found that the pipeline composed of KeyNetAffNetHardNet+SuperPoint is much better than various other combinations and singles, and achieved good results in the private list. It also achieved good results in the private list, see Table I.

\begin{table}[!ht]
    \centering
    \caption{The keypoint feature advancer and keypoint matcher of this project are much better than other open source popular ones, which set the number of keypoints to 8081, the keypoint detection threshold to 0.00102, and the image size to 1024.}
    \footnotesize
    \begin{tabular}{|l|l|l|l|}
    \hline
        Features & Matchers & Public & Private  \\ \hline
        ALIKED & LG & 0.141 & 0.157  \\ \hline
        Doghardnet & LG & 0.103 & 0.123  \\ \hline
        SuperPoint & LG & 0.092 & 0.108  \\ \hline
        DISK & LG & 0.092 & 0.118  \\ \hline
        ALIKED+DISK & LG+LG & 0.109 & 0.139  \\ \hline
        SIFT & LG & 0.114 & 0.118  \\ \hline
        Doghardnet+SIFT & LG+LG & 0.086 & 0.103  \\ \hline
        ALIKED+DISK+SIFT & LG+LG+LG & 0.140 & 0.141  \\ \hline
        SuperPoint+DISK+SIFT & LG+LG+LG & 0.153 & 0.155  \\ \hline
        DISK+SIFT & LG+LG & 0.134 & 0.136  \\ \hline
        SuperP oint+DISK+ALIKED & LG+LG+LG & 0.133 & 0.148  \\ \hline
        SuperPoint+DISK & LG+LG & 0.125 & 0.142  \\ \hline
        SuperPoint & SG & 0.112 & 0.134  \\ \hline
        KeyNetAffNetHardNet+SuperPoint (our) & AdaLAM+SG (our) & 0.177 & 0.167  \\ \hline
    \end{tabular}
\end{table}

\begin{table}[!ht]
    \centering
    \caption{The table is a comparative experimental study of SuperGlue, LightGlue, it is clear that LG is more suitable and advantageous than SG.}
    \footnotesize
    \begin{tabular}{|l|l|l|l|}
    \hline
        Features & Matchers & Public & Private  \\ \hline
        SuperPoint & LG & 0.092 & 0.108  \\ \hline
        SuperPoint & SG & 0.112 & 0.134  \\ \hline
    \end{tabular}
\end{table}

\subsection{Comparative Experiments of SuperGlue, LightGlue}

The project evaluates the performance difference between LG and SG in terms of both runtime and accuracy. Table 2 shows a comparison of the scores of LG and SG on the public and private test sets, and it is clear that LG is much better than SG.

\subsection{Experimental Results on Keypoint Feature Extraction and 3D Modeling}

This project uses KeyNetAffNetHardNet + AdaLAM and SuperPoint + SuperGlue two groups of keypoint extraction and keypoint matching methods, Figure 3 is the visualisation of KeyNetAffNetHardNet and SuperPoint two kinds of keypoint feature extraction, where the maximum number of keypoints is 4096. Figure 3 is the visualisation of KeyNetAffNetHardNet and SuperPoint keypoint feature extraction, where the maximum number of keypoints is 4096, the detection threshold is 0.000, the resolution of the image is 1024, and the effect of ensemble of the two methods, which is obviously much better than a single one.Figure 4 is the result of keypoint matching after ensemble of the two methods. Figure 5 is the 3D reconstruction of the whole pipeline and the result after ensemble.

\begin{figure}[!h]
    \centering
    \includegraphics[width=0.8\textwidth]{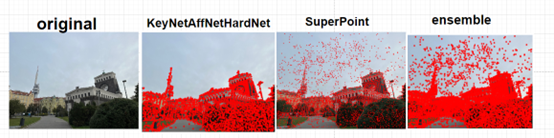}
    \caption{Keypoint feature extraction}   
    \label{fig4}
\end{figure}

\begin{figure}[!h]
    \centering
    \includegraphics[width=0.7\textwidth]{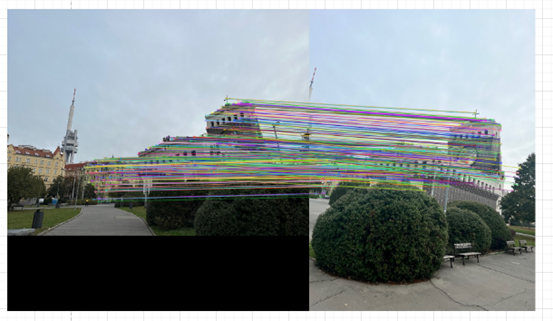}
    \caption{Key point matching}   
    \label{fig5}
\end{figure}

\begin{figure}[!h]
    \centering
    \includegraphics[width=0.7\textwidth]{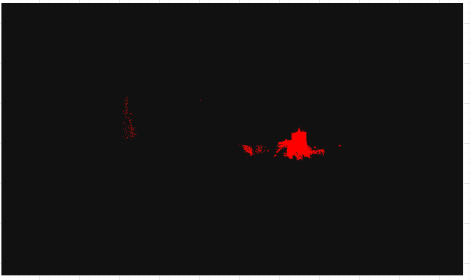}
    \caption{3D modelling}   
    \label{fig6}
\end{figure}

\section{Other Attempts}
\subsection{Other feature extractors and matchers}
The project uses one-stage dense matchers that have become popular in recent years, such as LoFTR [33], DKM [34], and the results are much worse, see Table III.

\begin{table}[!ht]
    \centering
    \caption{Experimental results of one-stage matcher}
    \begin{tabular}{|l|l|l|}
    \hline
        Phase 1 Matchers & Public & Private  \\ \hline
        LoFTR & 0.099 & 0.134  \\ \hline
        DKM & 0.071 & 0.084  \\ \hline
    \end{tabular}
\end{table}

\subsection{Image Rotation}
\begin{table}[!ht]
    \centering
    \caption{which shows comparative experiments with and without the use of image rotation}
    \begin{tabular}{|l|l|l|l|l|}
    \hline
        Features & Matchers & \makecell[c]{Whether or not the \\ image is rotated} & Public & Private  \\ \hline
        SP & LG & ~ & 0.092 & 0.108  \\ \hline
        SP & LG & $\checkmark$ & 0.069 & 0.091  \\ \hline
    \end{tabular}
\end{table}

In this part of the image rotation the project is using a pre-trained NetVLAD vision converter [38] to rotate the image when needed and then rotate its keypoints back to the original before starting the spatial position estimation, but it can be noticed by the above Table 4 that the results without the image rotation method are much better

\section{Acknowledgements}
We would like to thank the kaggle organisers of the image matching challenge and all the related organisations for their support and collaboration. This challenge not only helped the team to make technical progress and achieve good results, but also inspired the exploration and research of future image matching and 3D reconstruction techniques.



\nocite{*}
\bibliographystyle{unsrt} 
\bibliography{cas-refs}





\end{document}